\pdfoutput=1
\documentclass[11pt]{article}

\usepackage[]{EMNLP2023}

\usepackage{times}
\usepackage{latexsym}
\usepackage{amssymb}

\usepackage{times}
\usepackage{latexsym}
\usepackage[T1]{fontenc}
\usepackage[utf8]{inputenc}
\usepackage{subfigure}
\usepackage{microtype}
\usepackage{inconsolata}
\usepackage{amsfonts}
\usepackage{arydshln}

\usepackage{subfigure}
\usepackage{multirow}
\usepackage{graphicx} 
\usepackage{float}
\usepackage{CJKutf8}
\usepackage{amsmath}
\usepackage{stmaryrd}

\usepackage[T1]{fontenc}
\usepackage{graphicx}
\usepackage[utf8]{inputenc}

\usepackage{microtype}

\usepackage{inconsolata}

\usepackage{enumitem}
\usepackage{amssymb}

\title{UvA-MT's Participation in the WMT23 General Translation Shared Task}

\author{Di Wu* \qquad Shaomu Tan* \qquad David Stap \qquad Ali Araabi \qquad Christof Monz\\
  Language Technology Lab\\ University of Amsterdam \\
  \texttt{\{d.wu, s.tan, d.stap, a.araabi, c.monz\}@uva.nl} \\}

\begin{document}
\maketitle

\begin{abstract}
This paper describes the UvA-MT's submission to the WMT 2023 shared task on general machine translation. We participate in the constrained track in two directions: English $\leftrightarrow$ Hebrew. In this competition, we show that by using one model to handle bidirectional tasks, as a minimal setting of Multilingual Machine Translation (MMT), it is possible to achieve comparable results with that of traditional bilingual translation for both directions. By including effective strategies, like back-translation, re-parameterized embedding table, and task-oriented fine-tuning, we obtained competitive final results in the automatic evaluation for both English $\rightarrow$ Hebrew and Hebrew $\rightarrow$ English directions.
\end{abstract}

\def\thefootnote{*}\footnotetext{Equal contribution.}

\section{Introduction}
Multilingual Machine Translation~(MMT)~\citep{johnson-etal-2017-googles} has attracted a lot of attention in recent years because of 1) its high-level efficiency (multiple translation directions within a single model) and 2) the potential for knowledge transfer, especially for low-resource or even unseen directions. In MMT systems, only one additional tag is introduced to indicate the translation direction, compared to the conventional encoder-decoder architecture. In this competition, we explore MMT with a minimal setting, i.e., using one model for bidirectional translations simultaneously.

We leverage all the official parallel data and a substantial portion of the monolingual data generously provided by the WMT23 organizer, as elaborated in Section~\ref{data}. To elevate the quality of our parallel data, we implemented a comprehensive three-step cleaning procedure. Additionally, for monolingual data, we further trained an n-gram language model to filter out low-quality sentences, with the goal of generating synthetic data as elaborated in Section~\ref{pretrianing}.

The backbone of our system is based on a standard transformer~\citep{vaswani2017attention}.
Additionally, we build a re-parameterized embedding table ~\citep{wu2023beyond} (see Section~\ref{graphmerge}) to enhance the representational word similarity between English and Hebrew, targeting better knowledge transfer for multilingual translation.

The final system involves three stages of training: 1) \textbf{Pretraining with synthetic data} (see Section-\ref{pretrianing}), where we leverage back-translation~\citep{sennrich-etal-2016-improving} to produce synthetic data, and finally add them as additional data within new translation directions in our MMT system to conduct pretraining. 2) \textbf{Training without synthetic data} (see Section~\ref{training}), where we discard the additional synthetic data and further train our system using real bitext only. 3) \textbf{Fine-tuning with task-related data} (see Section~\ref{finetuning}), where we copy and fine-tune our system using English $\rightarrow$ Hebrew and Hebrew $\rightarrow$ English data for each track respectively. We observe evident improvements for stage 1 and stage 2, while surprisingly performance drops for stage 3.

We report our results, including the offline evaluation and the final online evaluation, in 
Section~\ref{results}. Our \textit{constrained} system showed comparable performance to \textit{unconstrained} systems, and outperformed the second-place constrained submission with +10 BLEU.

\section{Data} \label{data}

In this section, we provide an overview of our data sources and the data cleaning procedures applied to our English-Hebrew translation task. We utilize both parallel and monolingual data sets provided by the organizers for training our translation systems.

\subsection{Parallel Data}

We make use of all the available data from the constrained track of the shared task for English-Hebrew translation. To enhance the quality of our parallel data, we undergo a thorough preprocessing phase involving three key steps, as outlined below. All steps in the cleaning step 1 are executed using the Moses toolkit\footnote{\url{https://github.com/moses-smt/mosesdecoder/}} \cite{koehn2007moses}. Consequently, we reduced the size of the raw bitext data from 70 million to 34 million sentences after completing the three steps of the cleaning process:

\begin{itemize}[leftmargin=*]
   \item Cleaning Step 1
   \begin{itemize}
     \item Deescaping special characters in XML.
     \item Removing non-printable characters.
     \item Normalizing punctuation and tokenizing sentences using Moses.
   \end{itemize}
   \item Cleaning Step 2
   \begin{itemize}
     \item Filtering out sentences longer than 256 tokens.
     \item Eliminating sentences where over 75\% of the words on both the source and target sides are identical.
     \item Removing sentences with a source-to-target token ratio exceeding 1.5.
     \item Eliminating duplicate sentences.
   \end{itemize}
   \item Cleaning Step 3
   \begin{itemize}
     \item Removing off-target sentences using the FastText Language identification tool \cite{joulin2016fasttext}.
     \item Excluding sentences exhibiting one-to-many or many-to-one mappings, for example, a single source sentence having multiple different target sentences.
   \end{itemize}
 \end{itemize}

Furthermore, we sampled 10 million parallel sentences to learn a 32k joint unigram \cite{kudo2018subword} model-based subword vocabulary using SentencePiece \cite{kudo2018sentencepiece}, which we then utilized across all our models, including the n-gram KenLM model discussed in the next section. However, we encountered a situation where certain emoji tokens were not included in our vocabulary. As a result, we integrated an additional post-processing step in Section \ref{post-processing} to address this issue.

\subsection{Monolingual Data}
To enhance our translation systems further, we incorporate monolingual data to produce synthetic data through back-translation. For our monolingual data, we primarily rely on the official English data provided by the organizers. Note that we did not use any Hebrew monolingual data since it is limited (only 1 million sentences). We combine three official English monolingual datasets: News Discussions 2019, Leipzig News Corpora 2020, and News Crawl (2007-2022) to construct our raw monolingual dataset. Following this, we apply the same Cleaning Step 1 procedure as detailed in the Parallel Data section to preprocess the monolingual data, and this results in 373 million sentences. 

Considering the low quality of monolingual data, we additionally filter them by training an n-gram language model, i.e., KenLM~\citep{heafield2011kenlm}, and eliminate the sentences below an LM score threshold. The training data of KenLM is all of the test data in English, including our offline test data Flores, and the official test dataset. We train KenLM at the subword level, where we use the same unigram model (trained upon original bilingual data) to split the training data of KenLM. Then, we use it to score all of the monolingual data. To establish a filtering threshold, we randomly selected 1,000 sentences and labeled them as positive or negative based on criteria such as fluency, naturalness (e.g., avoiding strings of numbers), and relevance to the domain mentioned for WMT23 test datasets. Finally, we chose a threshold that could filter 70\% bad cases within the 1,000 sentences, with the cost of monolingual 30\% data, resulting in around 250M total sentences.

Lastly, considering the limitation of the computational resource, we sample 32M monolingual sentences (at the same level as the bilingual dataset) from the filtered dataset.

\section{Systems}

\subsection{Backbone and Baseline} \label{backbone}
In this section, we outline the foundational architecture and adjustments made to our baseline systems. Our baseline model leverages English $\leftrightarrow$ Hebrew translation directions by incorporating the target language token at the beginning of the encoder, denoted as "2he" and "2en". Our implementations are grounded in the Transformer architecture \cite{vaswani2017attention}, leveraging the Fairseq toolkit \cite{ott2019fairseq}.

For our baseline model, we utilize a 12-layer Transformer architecture (mT-large) with specific modifications, including pre-norm for both the encoder and decoder, and layer-norm for embedding. To enhance stability and performance, we tie the parameters of encoder embedding, decoder embedding, and decoder output. We also introduce dropout and attention dropout with a probability of 0.1, along with label smoothing at a rate of 0.1.

Similar to the approach described by \citet{vaswani2017attention}, we employ the Adam optimizer with a learning rate of 5e-4, implementing an inverse square root learning rate schedule with 4,000 warmup steps. We set the maximum number of tokens to 10,240, with gradient accumulation every 21 steps to facilitate large-batch training in \citet{tang2021multilingual}. We train all of our systems with 4 NVIDIA A6000 Gpus, and to expedite the training process, we conducted all experiments using half-precision training (FP16). Additionally, we save checkpoints every 2000 steps and implement early stopping based on perplexity, with a patience of 5 epochs.

\subsection{Re-parameterized Embedding Table} \label{graphmerge}
Using a vocabulary that is shared across languages is common practice in MMT. In addition to its simple design, shared tokens play an important role in positive knowledge transfer, assuming that shared tokens refer to similar meanings across languages. This point has been demonstrated by previous works~\citep{pires-etal-2019-multilingual,sun-etal-2022-alternative,stap2023viewing,wu2023beyond}. To enhance word-level knowledge transfer, we follow ~\citep{wu2023beyond} to implement a re-parameterized shared embedding table and equipped it with our backbone. 

We leverage eflomal~\citep{ostling2016efficient} to train and extract subword-level alignments based on all of the bilingual data we used. Then, we build the priors of word equivalence (word alignments) into a graph and leverage GNN~\citep{welling2016semi} to re-parameterize the embedding table. 

More specifically, For two words $v_{i}$ and $v_j$  in $V$, we define an alignment probability from $v_j$ to $v_i$ in corpus $D$ as corresponding transfer ratios $g_{i,j}$ as follows:

\begin{equation} \label{eq:03}
    g_{i,j}=\frac{c_{i,j}}{\sum_{k=1}^{|V|} c_{i,k}},
\end{equation}

\noindent%
where $c_{i,j}$ is the number of times both words are aligned with each other across $D$. The corresponding bilingual equivalence graph $G$ can be induced by filling an adjacency matrix using $g_{i,j}$, $G$ is applied within graph networks to re-parameterize the original embedding table as follows:

\begin{equation} \label{eq:07}
    E'= \rho (E W_{1} + G E W_{2}  + B).
\end{equation}

To allow the message to pass over multiple hops, we stack multiple graph networks and calculate representations recursively as follows:

\begin{equation} \label{eq:08}
    E^{h+1}= \rho (E^{h} W^{h}_{1} + G E^{h} W^{h}_{2}  + B^{h}),
\end{equation}

\noindent where $h$ is the layer index, i.e., \emph{hop}, and $E^{0}$ is equal to the original embedding table $E$. 
The last layer representation $E^{H}$ is the final re-parameterized embedding table, for the maximum number of hops $H$, which is then used by the system just like any vanilla embedding table. 

\begin{table*}[tp]
\centering
\scalebox{0.85}{
\begin{tabular}{l|cc|cc}
\hline\hline
\multirow{2}{*}{\textbf{Strategy}} & \multicolumn{2}{c|}{\textbf{Sampled Data}} & \multicolumn{2}{c}{\textbf{Full Data}}  \\
\cline{2-5} & \textbf{EN$\shortrightarrow$HE } & \textbf{HE$\shortrightarrow$EN } & \textbf{EN$\shortrightarrow$HE } & \textbf{HE$\shortrightarrow$EN } \\
\hline
Bilingual Baseline                           & 24.6 & 31.1 & 34.1 & 46.0 \\
MMT Baseline                                 & 24.7 & 31.6 & 34.1 & 45.8 \\
\hdashline
MMT + GM 1-hop                               &\bf 26.2 & 32.3 &\bf 34.3 &\bf 46.2 \\
MMT + GM 2-hop                               & 25.5 &\bf 32.7 & -	  & -    \\
\hline\hline
\end{tabular}
}
\caption{\label{table-1} Offline evaluation results on sampled and full training data. For sampled data (2M), the backbone is Transformer Base, while for full data (34M) the backbone is Transformer Large as we describe in Section~\ref{backbone}. MMT + GM means that we equip graph-based re-parameterized embedding tables for our MMT baseline, and hop means how many graph network layers are involved. The best BLEU scores in each column are written in bold.}
\end{table*}

\begin{table*}[tp]
\centering
\scalebox{0.9}{
\begin{tabular}{l|cc|cc}
\hline\hline
\multirow{2}{*}{\textbf{Strategy}} & \multicolumn{2}{c|}{\textbf{Offline}} & \multicolumn{2}{c}{\textbf{Online}}  \\
\cline{2-5} & \textbf{EN$\shortrightarrow$HE } & \textbf{HE$\shortrightarrow$EN } & \textbf{EN$\shortrightarrow$HE } & \textbf{HE$\shortrightarrow$EN } \\
\hline                                      
MMT Baseline                                 & 34.1    & 45.8   & 33.3 & 50.3    \\
\hdashline
MMT + GM 1-hop                                     & 34.3    & 46.2   & 33.6 & 50.7    \\
MMT + GM 1-hop + Stage-1                           &\bf35.4  & 46.8   &\bf 35.0 & 50.1    \\
MMT + GM 1-hop + Stage-1,2                         & 34.1    &\bf47.4 &\bf 35.0 &\bf 51.0    \\
MMT + GM 1-hop + Stage-1,2,3                       & 33.3    & 44.3   & 33.3 & 48.0    \\
\hline\hline
\end{tabular}
}
\caption{\label{table-2} Final results of three stages training. The best BLEU scores in each column are written in bold.}
\end{table*} 

\section{Experiments} \label{experiments}

We describe the training process of our system in three stages.

\subsection{Pretraining with Synthetic Data} \label{pretrianing}
Back-translation plays an important role in leveraging monolingual data in machine translation. In this competition, we also apply it to produce synthetic data and include it in our first-stage training. 

Specifically, we first train a base MMT model (backbone with re-parameterized embedding tables) using bilingual data. Then, we feed our monolingual English data to produce EN-HE synthetic bitext. Finally, we merge the original bilingual data with the synthetic data together to pre-train our MMT system. We follow ~\citet{fan2021beyond} and add an additional language tag "2syn" to differentiate between synthetic and original Hebrew data. Note that, although normally original data (here, it is EN) is used as target side data after back translation, we use synthetic data for both directions.

\subsection{Training without Synthetic Data} \label{training}
Considering that the synthetic data may differ from the original bilingual data in terms of data quality, domain difference, and diversity, in the second-stage training, we encourage our system to skew towards the original bilingual distribution. We achieve this by discarding the synthetic data directly and continuing training upon the first-stage system directly as a kind of full parameter finetuning. 

\subsection{Finetuning on Task-Specific Data} \label{finetuning}
Lastly, to encourage the system to focus on one certain language direction, we further fine-tune direction-specific data on the second-stage system. Note, the direction-specific data here, i.e., EN $\rightarrow$ HE and HE $\rightarrow$ EN are both from the original bilingual data. The effectiveness is also demonstrated by ~\citet{ding-etal-2021-improving,zan-etal-2022-vega} for bilingual translation. 

In short, in this three-stage training process, we gradually narrow down the data distribution to focus on task-specific real data.

\subsection{Post-Processing} \label{post-processing}

We noticed that some emoji tokens in the official test set were not included in our vocabulary. Thus, we integrated an additional post-processing step to process them. Specifically, we replaced the emoji tokens with their Unicodes before feeding them to our system to conduct inference, and then convert the Unicodes back for generated predictions.

\subsection{Offline Evaluation} \label{evaluation}

We used Flores-200~\citep{costa2022no} to evaluate our strategies offline before submissions and Ntrex-128~\cite{federmann2022ntrex} as the validation set. We show the results in Table~\ref{table-1}. Due to resource limitations, we sample 2M of bilingual data to verify whether there is a big performance gap between MMT and bilingual baseline. Meanwhile, we also chose the best hyperparameter for our re-parameterized embedding table, i.e., the hop number, based on the sampled dataset.

As shown in Table~\ref{table-1}, on both sampled and full data, the MMT baseline achieves comparable results with bilingual counterparts. Especially for sampled data, it even outperforms 0.5 BLEU for into-English translation. 

The model-equipped 1-hop re-parameterized embedding table demonstrates a notable improvement, yielding a 1.5 BLEU gain for the out-of-English direction and a 0.7 BLEU gain for the out-of- and into-English directions, on 2M datasets. It shows that the embedding re-parameterized method~\cite {wu2023beyond} also works for bilingual settings, which is not explored in the original paper. We did not observe evident gains for 2-hop compared with 1-hop on sampled data, hence, we only apply the 1-hop graph networks for the full data training. As shown in the table, the results are consistent with that of small data, where MMT with 1-hop graph networks achieve better performance than MMT baseline.

As above, we chose MMT with the 1-hop setting as our architecture and conducted our three stages of training as described in Section~\ref{experiments}.

\section{Results} \label{results}
Table-\ref{table-2} shows our offline and online evaluation results according to each training stage described in Section~\ref{experiments}. We still use Flores-200 to conduct offline evaluations. The online results are reported by WMT23 background BLEU evaluations. Stage-1, -2, and -3 refer to "Pretraining with Synthetic Data", "Training without Synthetic Data", and "Finetuning on Task-Specific Data" respectively.

The results of online and offline evaluations are quite consistent. Both of them achieve best results when training with stage 1 and stage 2. It shows that by step-by-step narrowing training data from mixing with synthetic data to real data distribution, we can further boost our MMT system's performance. However, when we further conduct fine-tuning on direction-specific data, i.e., applying stage 3, there is an evident performance drop. It seems that tuning in a specific direction upon MMT may not be a good practice, at least when the training data are a subset of that for MMT. We leave this point for future exploration.

Our final system achieves 35.0 and 51.0 in EN$\shortrightarrow$HE and HE$\shortrightarrow$EN direction respectively, which are both in the first place for constrained tracks.

\section{Conclusion}
In this competition, we show that: 1) It is possible to achieve comparable results with conventional bilingual translation by using MMT training fashion to handle two dual translation directions. 2) Previous embedding re-parameterized method~\citep{wu2023beyond} also works for bilingual translation, which is not verified in the original paper. However, when training data scales up to 30+M level, the improvements become marginal. 3) By step-by-step narrowing training data (especially for stage-1 and stage-2) from mixing with synthetic data to real data distribution, we successfully boost the final performance, even in a quite high-resource scenario (30+M).

\bibliography{anthology,custom}
\bibliographystyle{acl_natbib}

\appendix

\end{document}